\documentclass[12pt,a4paper,twoside, twocolumn]{article}
\usepackage{dhasa}

\usepackage{graphicx}
\usepackage{epsfig}
\usepackage{amssymb}
\usepackage{amsmath}
\usepackage{amsfonts}
\usepackage{color}
\usepackage{adjustbox}
\usepackage{algorithm}
\usepackage{graphicx}

\begin{document}

\title*{Izindaba-Tindzaba: Machine learning news categorisation for Long and Short Text for isiZulu and Siswati}
\textit{Madodonga, Andani}\\
\textit{Department of Computer Science, University of Pretoria, South Africa} \\
\textit{andanim412@gmail.com} 

\textit{Marivate, Vukosi }\\
\textit{Department of Computer Science, University of Pretoria, South Africa} \\
\textit{vukosi.marivate@cs.up.ac.za} 

\textit{Adendorff, Matthew}\\
\textit{Open Cities Lab} \\
\textit{matthew@opencitieslab.org} 

\section*{Abstract}
Local/Native South African languages are classified as low-resource languages. As such, it is essential to build the resources for these languages so that they can benefit from advances in the field of natural language processing. In this work, the focus was to create annotated news datasets for the isiZulu and Siswati native languages based on news topic classification tasks and present the findings from these baseline classification models. Due to the shortage of data for these native South African languages, the datasets that were created were augmented and oversampled to increase data size and overcome class classification imbalance. In total, four different classification models were used namely Logistic regression, Naive bayes, XGBoost and LSTM. These models were trained on three different word embeddings namely Bag-Of-Words, TFIDF and Word2vec. The results of this study showed that XGBoost, Logistic Regression and LSTM, trained from Word2vec performed better than the other combinations.


Keywords: South African native Languages, Low Resources Languages, Data Augmentation, Topic Classification, News Categorisation

\section{Introduction}
Natural Language Processing (NLP) is a subfield of artificial intelligence, linguistics and computer science that focuses on enablling computers to process natural language \citep{artificialintelligencebigdataanalyticsandinsight_2020}. One of the cases where NLP has been beneficial to people is where it has been used for machine translation, performing the task of translating from one language to another. In this case, NLP helps the computer or machine to attempt the conversion from one langauge to another. NLP can also assist in learning and prediction sentiment/opinion from sentences or text. This NLP capability is utilised by companies to understand how customers feel and their opinion about the company’s products and services through the analysis of their social media posts and comments. Furthermore, the chatbots that are used in the customer services space are one of the examples of NLP application 
\citep{artificialintelligencebigdataanalyticsandinsight_2020}. Contextual chatbots and Virtual Text Assistant are now widely used but they mostly understand a limited number of languages, such as English. South African native languages do not have enough resources to be used to built such contextual Chatbots and Virtual Text Assistant. Therefore, the resources for native languages need to be created so that they can be used to build software agents that understand South African native languages \citep{duvenhage2017improved}.

South Africa is a multilingual country with eleven langauges (two of which are European and nine are African languages); the African languages are Sepedi, Sesotho, Setswana, Siswati, Tshivenda, Xitsonga, isiZulu, isiNdebele and isiXhosa and on the other hand, European languages are English and Afrikaans. It is important to note that these languages are official in South Africa \citep{gateway_2021}. In South Africa, we have a challenge with the nine African languages because they are resource-poor. There is a shortage of curated and annotated corpora to enable them to benefit from Natural Language Processing. Therefore, the purpose of this study is to focus specifically on the corpus creation and annotation for isiZulu and Siswati and perform a topic classification tasks on the data.

\section{Critical Natural Language Processing Components}
\label{sec:first:first_sec1}

Globalisation and the increase in digital communications have created the demand for NLP systems that enable fast communication between people speaking different languages. However, some languages are missing in these systems. For instance, there are roughly 7000 spoken languages on the planet and Most of them still are not included in the NLP systems, primarily because they do not have the labelled corpora to build those NLP systems \citep{baumann2014using}. These languages with scarce or no resources are low-resourced languages \citep{whyatt2019languages}. The language resources include (but are not limited to) the annotated corpora and core technologies. Examples of core technologies include lemmatisers, part of speech tagger and morphological decomposers  \citep{eiselen2014developing}.
 On the other hand, the languages with high resources are the ones that have most of the resources needed to build the NLP technology \citep{xu2013cross}.

The high-resourced languages include English, French, Finnish, Italian, German, Mandarin, Japanese, etc.
\citep{bonab2019simulating,xu2013cross} and low-resourced languages include languages such as isiZulu, isiXhosa, Siswati etc. \citep{bosch2008experimental}. A study, by \cite{eiselen2014developing}, focused on the low-resourced languages, namely, isiZulu and Siswati; stated that annotated corpora are one of the things that low-resourced languages lack. Thus, the isiZulu and Siswati datasets need to be annotated, as part of the process of making these languages accessible for NLP and by enriching these two languages. \cite{hsueh2009data} defines data annotation as the process of labelling the dataset(s), an important step when building machine learning models. \cite{stenetorp2012brat} stated that manual data annotation is the most important, time-consuming, costly, and tedious task for NLP researchers. Therefore, automation tools are developed to perform these annotations. 

The lack of curated and annotated data impede the process of fighting the shortage of resources for low-resourced languages in the NLP space \citep{niyongabo2020kinnews}. Besides, established NLP methods often cannot be transferred on or to these languages without these corpora \citep{niyongabo2020kinnews}. 
 \cite{niyongabo2020kinnews} collected the datasets of two closely related African languages - Kirundi and Kinyarwanda from two different sources. A total of 21268 and 4612 articles were annotated for Kinyarwanda and Kirundi respectively. The two datasets underwent a cleaning process that involved the removal of special language characters and stopwords. 
The sources were newspapers and websites. These datasets were annotated, based on the title and content of the contained articles, into the following categories:\textit{ Politics; Sport; Economy; Health, Entertainment; History; Technology; Tourism; Culture; Fashion; Religion, Environment; Education; and Relationship} \citep{rakholia2016lexical}. Hence, a very similar task was performed in this work as part of language resources creation.

\subsection{Data generation techniques for low-resourced languages}
\label{sec:first:first_sec3}

An existing approach utilised to mitigate the challenges of low-resourced language data, is the language translation approach. That is the low-resourced language gets translated into the resource-rich language \citep{tang2018improving}. However, in most cases, this approach suffers from language biases and may be impractical to achieve in real life \citep{tang2018improving}. Sometimes the direct translation may be impossible or inaccurate due to language differences. Hence, the translated data will require manual processing thereafter, which is tedious and time-consuming. Manually creating data for low-resourced languages is time-consuming but a good approach, moreover, it introduces minimal language biases and more accurate than translated datasets \citep{shamsfardchallenges}.

Cross-lingual and transfer learning is one of the combinations of techniques frequently used or preferred in NLP due to its speed and efficiency \citep{shamsfardchallenges}. This further serves to highlight why all languages must have NLP resources such as annotated data to avoid data simulations that have unfavourable
effects.

Data Augmentation is a method that generates a copy (or unique data) of the data by slightly altering the existing data \citep{duong2021review}. It increases the size of small training data in ways that improve model performance \citep{abonizio2020pre}. Model performance is highly dependent on the quality and size of the training data. Data Augmentation addresses the issue of small training data that leads to the models losing their generalisability \citep{kobayashi2018contextual}.

Work by \cite{marivate2020investigating} had a small data size of Sepedi and Setswana native languages, and incorporated word embeddings based-contextual augmentation to increase the dataset used to train classification models. Each training dataset was augmented 20 times while the test dataset remains unchanged.
In their study, the new data created replaced the words (based on context) in the sentences. Hence a new sentence was formed as a result of applying Contextual Data Augmentation. Furthermore, Data Augmentation improved the performance of the classifiers \citep{marivate2020investigating}. In this current study, the same Data Augmentation (word embedding-based augmentation) was performed on the Siswati and isiZulu dataset to increase the data size. 

\subsection{Dealing with data imbalance} 

The Synthetic Minority Oversampling Technique (SMOTE)  is another technique that can be  adopted when the learning is done on an imbalanced dataset, since it solves the problem of class imbalance \citep{fernandez2018smote}. SMOTE works by generating synthetic examples through inserting different values(words) in minority class, the values are randomly picked from a defined neighbourhood within feature space . Minority class is selected, then obtain the k-nearest neighbours of the same minority class and therefore utilises the k- neighbours to create the new synthetic examples \citep{fernandez2018smote}.

\subsection{Related work}
Supervised learning models perform better on larger labelled datasets, which presents a challenge for low-resourced languages as they don’t have enough data and annotating data can be expensive \citep{fang2017model}. Most prior studies focused on developing parallel corpora between low and  resource-rich languages, but parallel corpora are often unavailable for some low-resourced languages \citep{fang2017model}. Work by \cite{zoph2016transfer} identified low-resourced languages and investigated the idea of distance learning on machine translation. 
Since English and French are resource-rich languages, the two languages trained a neural machine translation (NMT) \citep{zoph2016transfer}. 
An English-French neural machine translation (NMT) model was initially trained. Afterwards, the NMT model initialised another NMT model to be used on a low-resourced and high-resourced pair (e.g. Uzbek-English) \citep{zoph2016transfer}, as such utilising transfer learning. In this case, the low-resourced languages investigated for transfer were Uzbek, Hausa, Turkish and Urdu. The transfer learning was shown to improve the BLEU (bilingual evaluation understudy) for low-resourced Neural machine translation \citep{zoph2016transfer}. 

Work by \cite{nguyen2017transfer} explored transfer learning between the two low-resourced languages Turkey and Uzbek by first pairing each language with English and then generating the parallel data. Then, split the words with Bytes Pair Encoding (BPE) to maximise the overlapping vocab \citep{nguyen2017transfer}. The model and word embedding are trained on the first language pair (Turkey-English) and then the same model parameters and word embeddings were transferred to the other model that trained the second language pair (Uzbek-English). This technique improved the BLEU by 4.3\% \citep{nguyen2017transfer}. 


The datasets of low-resourced South African languages, isiZulu collected from isolezwe and National Centre for Human Language Technology (\url{www.sadilar.org}); and Sepedi collected from National Centre for Human Language Technology were used to evaluate the performance of open-vocabulary models on the small datasets, the evaluated models include n-grams, LSTM, RNN, FFNN, and transformers. The performance of the models was evaluated using the byte pair encoding (BPE).  
 The RNN performed better than the rest of the models on both the isiZulu and Sepedi datasets \citep{mesham2021low}. \cite{nyoni2021low} explored the machine translation capability from the zero-short learning, transfer learning and multilingual learning on two South African languages, namely, isiZulu and isiXhosa; and one Zimbabwean language, that is Shona. The datasets were in language pair (parallel text), that is, English-to-Shona, English-to-Zulu, English-to-Xhosa and Zulu -to- Xhosa, with the pair English -to- Zulu being the target pair since it has the smallest datasets (sentence pair). The transfer learning and zero-short learning did not outperform the multilingual model which produced the Bleu score of 18.6 for the English-to-Zulu pair. Moreover, these results provide an avenue for the development and improvement of low resource translation techniques \citep{nyoni2021low}.
 
Work by \cite{marivate2020investigating} attempted to address  the issue of lack of clear guidelines for low-resources languages in terms of collecting and curating the data for specific use in the Natural Language Processing domain. In their investigation, two datasets of news headlines written in Sepedi and Setswana were collected, curated, annotated, and fed into the machine learning classification models to perform text classification. The datasets were annotated by means of categorising the articles into the following categories based on context: \textit{Legal; General News; Sports; Politics; Traffic News; Community Activities; Crime; Business; Foreign Affairs} \citep{marivate2020investigating}. The evaluation metric was the F1-score, which is a model performance measure. One of the models, Xgboost, performed well as compared to other models \citep{marivate2020investigating}.

\section{Developing news classification models for isiZulu and Siswati languages}

In this section we discuss data collection and cleaning processes together with the classification models building approach.

\subsection{Data Collection, Cleaning and Annotation}

We discuss the initial news data collection and annotation process. We further discuss the data collection process of the larger dataset that was used to build our word representations.

\subsubsection{News data collection and annotation}
The isiZulu news data was collected from Isolezwe, which is a Zulu-language local newspaper. The news articles published online on Isolezwe website (\url{http://www.isolezwe.co.za}) were scraped and stored in a csv file for further processing. The Siswati dataset (news headlines) was collected from the public broadcaster for South Africa, that is, SABC news LigwalagwalaFM Facebook page (\url{https://www.facebook.com/ligwalagwalafm/}). The Siswati data was also scraped and stored on a csv file. Lastly, to build word respresentations other isiZulu and Siswati datasets were collected from SADILAR (\url{www.sadilar.org}) and Leipzig Corpus(\url{https://wortschatz.uni-leipzig.de}) for the purpose of better generalising word representations. We collected 752 (full artilces and titles) in isiZulu and collected 80 Siswati news headlines.



Post data collection process, we worked to categorise the news items using the International Press Telecommunications Council (IPTC) News Categories (or codes)\footnote{\url{https://iptc.org/standards/newscodes/}}. The categories used were: \textit{1. disaster, accident and emergency incident, 2. economy, business and finance, 3. education, 4. environment, 5. health, 6. human interest, 7. labour, 8. lifestyle and leisure, 9. politics,	10. religion and belief, 11. science and technology, 12. society, 13. sport, 14. weather, 15. arts, culture, entertainment and media, 16. crime, law and justice, 17. conflict, war and peace}. We make available the data, and annotations and data statement\footnote{\url{https://github.com/dsfsi/za-isizulu-siswati-news-2022}} \footnote{\url{https://doi.org/10.5281/zenodo.7193346}}. An example of an annotated isiZulu article is shown below:

\textbf{Politics} \\
\textit{UMENGAMELI we-ANC uMnuz Cyril Ramaphosa ugqugquzele abantu basePort Shepstone nezindawo ezakhele leli dolobha ukuthi bagcwalise iMoses Mabhida Stadium lapho ezothula khona umyalezo wakhe weJanuary 8 ngoMgqibelo aphinde athule nomhlahlandela weqembu wokuheha abavoti njengoba kuyiwa okhethweni.
URamaphosa ehambisana nabanye abaholi be-ANC esifundazweni uhambele kule ndawo izolo enxenxa abantu ukuthi batheleke ngobuningi kulo mgubho weqembu. Uphinde wathembisa ukuthi uzothula uhlelo lwakhe lokuthuthukisa izwe.}

The isiZulu news (articles and titles) and Siswati news titles category distribution are shown below, it was observed that the datasets suffer from class imbalance, small data size and short text (only isiZulu and Siswati titles/headlines). Therefore, oversampling techniques, SMOTE and Data Augmentation were applied to mitigate class imbalance problem and also increase the data size.
\begin{figure}[H]
	\centering {\includegraphics[width=1\columnwidth]{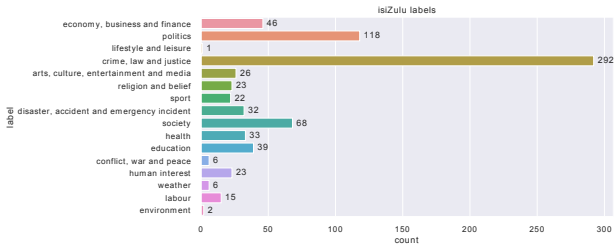}}
	\caption[isiZulu initial Class Distribution]{isiZulu initial Class Distribution}
	\label{fig:fig1ak}
\end{figure}

\begin{figure}[H]
	\centering {\includegraphics[width=1\columnwidth]{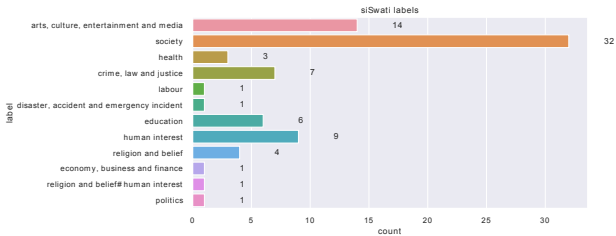}}
	\caption[Siswati initial Class Distribution]{Siswati initial Class Distribution}
	\label{fig:fig1bk}
\end{figure}

For better modelling, class categories with few observations we revmoved, remaining with the below categories:
\textit{1. crime, law and justice, 2. economy, business and finance, 3. education, 4. politics, 5. society}  for isiZulu  and \textit{1. crime, law and justice, 2. arts,culture,entertainment and media, 3. education, 4. human interest, 5. society} for Siswati. Since the number of class categories has dropped to 5 categories, the news dataset size also dropped to 563 (news articles and titles) for isiZulu and 68 (news titles) for Siswati 
The final datasets were cleaned and then used to build classification models, however, prior to model building, word representations were created using a larger datasets. 

\subsection{Data Preparation/Cleaning}
All datasets collected in this work contained some noise such as single characters, white spaces, encoded characters, meaningless words, and special characters. The noise had to be removed before the datasets are fed into the models. All these noises on the datasets were removed. Below we explain each part of the followed cleaning step: 

\begin{itemize}
\item The single characters carry less meaning, so they were removed from the datasets.
\item There were instances where there are multiple spaces between two words, so those spaces were substituted with a single space.
\item There were some characters/words that were not ASCII encoded then those characters were decoded back to ASCII.
\item Special characters refer characters such as \&\%\$ and they are not accepted by the models. Hence they were also removed.
\item The data contained combination of letters that don’t make any existing isiZulu/Siswati word. Words such as \textit{'udkt','unksz','unkk'}
\end{itemize}. Based on these criteria, they were also removed to streamline the corpus, and as a result, improve the analysis.

Since the datasets are noise-free, each letter in the datasets was set to lowercase, resulting in clean datasets to be used in machine learning models building.

\subsubsection{Word Representations}
It was stated above that the larger datasets collected from SADILAR and Leipzig Corpus for each language was used for word representations (vectorisers and embeddings) creation. The pre-trained vectorizers were created, enabling the opportunity to build classifiers with good generalisability in future. Therefore, from the collected corpora for each language, we created the following vectorizers: Bag Of Words, TFIDF and Word2vec \citep{NIPS2013_9aa42b31}.
 
 \begin{table}[H]
	\caption[Data collection]{Vectorizer Corpora Sizes in number of tokens}
	\centering
	\begin{tabular}{|l|ll|l}
\cline{1-3}
 & \multicolumn{2}{c|}{\textbf{Tokens}} &  \\ \cline{1-3}
 \textbf{Source} & \multicolumn{1}{l|}{\textbf{isiZulu}} & \textbf{Siswati}   \\ \cline{1-3}
Sadilar & \multicolumn{1}{l|}{770845} & 399800  \\ \cline{1-3}
Leipzig & \multicolumn{1}{l|}{4296659} & 134827   \\ \cline{1-3}
Total & \multicolumn{1}{l|}{5067504} & 534627   \\ \cline{1-3}
\end{tabular}
	\label{tab:table2}
\end{table}

\subsection{News Classification Models}
We arbitrarily selected a few classification algorithms to train models to perform news topic classification for isiZulu and Siswati datasets. The selected algorithm are Logistic Regression, XGBoost, Naive Bayes and LSTM. 

We performed the classification on the original datasets, and then apply oversampling techniques, namely, Data Augmentation and SMOTE, to solve the class imbalance problem and increase the data size. The classification models were again executed on the Augmented and SMOTE datasets.

\section{Experiments and Results}
In this section we discuss the results obtained from the performed experiments, that is, the findings from the multiple combination of word representations and classification models on isiZulu and Siswati datasets. However, the findings presented here are basis, since this work only provide guidelines for resource creation of low-resource languages. 

\subsection{Experimental Setup}
The maximum token size of 20 000 was used for both Bag Of Words and TFIDF vectorizers, whereas for Word2vec we  used size 300. For each of the 4 classification models, 5-fold cross validation was applied during model training. As we are creating baseline models and working on small datasets (not enough to split into training, validation and test sets), then parameter optimisation was not performed in this work. 

\subsubsection{Baseline Experiments} 
In the baseline experiments, we train the classification models using 5-fold cross validation on isiZulu and Siswati original datasets and present the models performance for each dataset.  The results show that Word2vec and LSTM model performed very well in all datasets as compared to other models. Below tables shows the classification model results obtained from original datasets.

\begin{table}[htb!]
\caption[isiZulu Articles Original Dataset Model Performance]{isiZulu Articles Original Dataset Model Performance}
\label{tab:table1a}
\centering
\resizebox{\linewidth}{1.5cm}{
\begin{tabular}{|c|c|c|c|c|c|c|}
\hline
\textbf{Preprocessing} & \textbf{Model}      & \textbf{Precision(\%)} & \textbf{Recall(\%)} & \textbf{F1-score(\%)} & \textbf{Accuracy(\%)} & \textbf{Confidence   Interval(f1 score)} \\ \hline
Bag-Of-Words        & Naive Bayes         & 21.73                  & 21.12               & 16.34                 & 52.4                  & (13.29,19.4)                             \\ \hline
Bag-Of-Words        & Logistic Regression & 41.23                  & 34.97               & 36.06                 & 54.53                 & (32.09,40.03)                            \\ \hline
Bag-Of-Words        & XGBoost             & 49.14                  & 31.33               & 32.51                 & 54.89                 & (28.64,36.38)                            \\ \hline
TF-IDF        & Naive Bayes         & 18.41                  & 20.34               & 14.35                 & 52.22                 & (11.45,17.24)                            \\ \hline
TF-IDF        & Logistic Regression & 32.09                  & 26.13               & 24.19                 & 54.71                 & (20.65,27.73)                            \\ \hline
TF-IDF        & XGBoost             & 40.91                  & 29.42               & 29.34                 & 52.93                 & (25.58,33.1)                             \\ \hline
Word2vec               & Naive Bayes         & 61.98                  & 50.99               & 53.04                 & 68.39                 & (48.91,57.16)                            \\ \hline
Word2vec               & Logistic Regression & 70.18                  & 62.91               & 65.13                 & 75.32                 & (61.19,69.07)                            \\ \hline
Word2vec               & XGBoost             & 67.69                  & 52.23               & 55.83                 & 69.1                  & (51.73,59.93)                            \\ \hline
Word2vec               & LSTM                & \textbf{83.39}         & \textbf{83.11}      & \textbf{82.78}        & \textbf{83.11}        & (79.66,85.9)                             \\ \hline
\end{tabular}}
\end{table}

\begin{table}[htb!]
\caption[isiZulu Titles Original Dataset Model Performance]{isiZulu Titles Original Dataset Model Performance}
\label{tab:table2a}
	\centering
	\resizebox{\linewidth}{1.5cm}{
\begin{tabular}{|c|c|c|c|c|c|c|}
\hline
\multicolumn{1}{|c|}{\textbf{Preprocessing}} & \multicolumn{1}{c|}{\textbf{Model}} & \multicolumn{1}{c|}{\textbf{Precision(\%)}} & \multicolumn{1}{c|}{\textbf{Recall(\%)}} & \multicolumn{1}{c|}{\textbf{F1-score(\%)}} & \multicolumn{1}{c|}{\textbf{Accuracy(\%)}} & \multicolumn{1}{c|}{\textbf{Confidence Interval(f1 score)}} \\ \hline
Bag-Of-Words                              & Naive   Bayes                       & 17.6                                        & 20.62                                    & 15.33                                      & 51.69                                      & (12.36,18.31)                                               \\ \hline
Bag-Of-Words                              & Logistic   Regression               & 18.36                                       & 21.83                                    & 17.38                                      & 52.76                                      & (14.25,20.51)                                               \\ \hline
Bag-Of-Words                              & XGBoost                             & 20.91                                       & 21.23                                    & 17.03                                      & 51.51                                      & (13.92,20.13)                                               \\ \hline
TF-IDF                              & Naive   Bayes                       & 19.89                                       & 20.89                                    & 15.57                                      & 52.4                                       & (12.57,18.56)                                               \\ \hline
TF-IDF                              & Logistic   Regression               & 20.47                                       & 21.9                                     & 17.58                                      & 52.93                                      & (14.44,20.73)                                               \\ \hline
TF-IDF                              & XGBoost                             & 18.07                                       & 20.79                                    & 16.37                                      & 51.34                                      & (13.31,19.43)                                               \\ \hline
Word2vec                                     & Naive   Bayes                       & 27.83                                       & 25.58                                    & 22.75                                      & 57.2                                       & (19.29,26.22)                                               \\ \hline
Word2vec                                     & Logistic   Regression               & 41.85                                       & 38.65                                    & 39.18                                      & 57.72                                      & (35.14,43.21)                                               \\ \hline
Word2vec                                     & XGBoost                             & 40.63                                       & 31.17                                    & 31.03                                      & 57.73                                      & (27.21,34.85)                                               \\ \hline
Word2vec                                     & LSTM                                & \textbf{72.96}                              & \textbf{71.75}                           & \textbf{72.01}                             & \textbf{71.75}                             & (68.3,75.72)                                                \\ \hline
\end{tabular}}

\end{table}

\begin{table}[htb!]
	\caption[Siswati Titles Original Dataset Model Performance]{Siswati Titles Original Dataset Model Performance}
	\centering 
	\resizebox{\linewidth}{1.5cm}{
	\begin{tabular}{|c|c|c|c|c|c|c|}
\hline
\multicolumn{1}{|c|}{\textbf{Preprocessing}} & \multicolumn{1}{c|}{\textbf{Model}} & \multicolumn{1}{c|}{\textbf{Precision(\%)}} & \multicolumn{1}{c|}{\textbf{Recall(\%)}} & \multicolumn{1}{c|}{\textbf{F1-score(\%)}} & \multicolumn{1}{c|}{\textbf{Accuracy(\%)}} & \multicolumn{1}{c|}{\textbf{Confidence   Interval(f1 score)}} \\ \hline
Bag-Of-Words                              & XGBoost                             & 25.75                                       & 25.52                                    & 24.23                                      & 41.54                                      & (14.05,34.42)                                                 \\ \hline
Bag-Of-Words                              & Naive   Bayes                       & 25.37                                       & 30                                       & 25.39                                      & 53.19                                      & (15.04,35.73)                                                 \\ \hline
Bag-Of-Words                              & Logistic   Regression               & 25.93                                       & 30.1                                     & 26.34                                      & 48.79                                      & (15.87,36.81)                                                 \\ \hline
TF-IDF                              & Naive   Bayes                       & 13.61                                       & 22                                       & 15.61                                      & 48.68                                      & (6.98,24.23)                                                  \\ \hline
TF-IDF                              & Logistic   Regression               & 17.77                                       & 24                                       & 18.81                                      & 50.33                                      & (9.52,28.1)                                                   \\ \hline
TF-IDF                              & XGBoost                             & 25.16                                       & 29.33                                    & 25.5                                       & 47.58                                      & (15.14,35.86)                                                 \\ \hline
Word2vec                                     & Naive   Bayes                       & 31.77                                       & 34.76                                    & 31.57                                      & 59.01                                      & (20.52,42.61)                                                 \\ \hline
Word2vec                                     & Logistic   Regression               & 29.59                                       & 32                                       & 28.09                                      & 57.58                                      & (17.4,38.77)                                                  \\ \hline
Word2vec                                     & XGBoost                             & 28.77                                       & 31.43                                    & 27.96                                      & 54.84                                      & (17.29,38.62)                                                 \\ \hline
Word2vec                                     & LSTM                                & \textbf{87.53}                              & \textbf{80.88}                           & \textbf{81.06}                             & \textbf{80.88}                             & (71.75,90.37)                                                 \\ \hline
\end{tabular}}
	

	\label{tab:fig15a11}
\end{table}

\subsubsection{Augmentation}
Data Augmentation is the technique that is used to increase the data size to improve the performance of the machine learning classifiers \cite{oh2020time}. The most common way to augment the data is by means of replacing the words or phrases in a sentence by their synonyms where the synonym is derived by obtaining the semantically close words \citep{zhang2015character}.

The Siswati and isiZulu datasets were augmented using the same approach where the original words on the sentence are replaced based on their contextual meaning. The augmentation was done through referencing the words similarity from the Word2vec word embedding as per \cite{marivate2020investigating}. Data Augmentation improved the performance of each model on all datasets as compared to original datasets, hence, it remains a task to investigate the effectiveness and robustness of this Data Augmentation algorithm, that can be achieved through comparing the algorithm results on resourced and low-resourced datasets.

The classification models trained on Word2vec outperformed all the classification models trained on TFIDF and Bag Of Words. For isiZulu articles, combination of Word2vec and XGBoost model outperformed all the models, scoring f1-score of 95.21\%, on the other hand, Word2vec and Logistic Regression model  combination performed well on isiZulu titles dataset scoring f1-score of 86.42\%. Lastly, Word2vec and  LSTM model combination performed well on Siswati titles dataset scoring f1-score of 93.15\%. It was observed that iziZulu articles dataset scored high f1-score as compared to isiZulu titles, which explains that long texts improves the classification accuracy, and also highlights that Logistic Regression outperforms XGBoost on short text dataset. It remains a task to run the same comparison on Siswati dataset, as it was not covered in this work due to lack of Siswati full news articles dataset. 



\begin{table}[htb!]
\caption[isiZulu Articles Augmented Dataset Model Performance]{isiZulu Articles Augmented Dataset Model Performance}
\label{tab:table3a}
\centering
\resizebox{\linewidth}{1.5cm}{
\begin{tabular}{|c|c|c|c|c|c|c|}
\hline
\textbf{Preprocessing} & \textbf{Model}        & \textbf{Precision(\%)} & \textbf{Recall(\%)} & \textbf{F1-score(\%)} & \textbf{Accuracy(\%)} & \textbf{Confidence Interval(f1 score)} \\ \hline
Bag-Of-Words        & Naive   Bayes         & 71.65                  & 68.55               & 68.42                 & 68.89                 & (65.87,70.97)                          \\ \hline
Bag-Of-Words        & Logistic   Regression & 83.35                  & 83.92               & 83.09                 & 83.23                 & (81.04,85.15)                          \\ \hline
Bag-Of-Words        & XGBoost               & 74.28                  & 73.85               & 73.68                 & 73.51                 & (71.26,76.09)                          \\ \hline
TF-IDF        & Naive   Bayes         & 75.71                  & 73.77               & 73.6                  & 73.98                 & (71.18,76.02)                          \\ \hline
TF-IDF        & Logistic   Regression & 79.65                  & 79.91               & 79.2                  & 79.39                 & (76.97,81.42)                          \\ \hline
TF-IDF        & XGBoost               & 80.44                  & 80.44               & 79.92                 & 80.02                 & (77.72,82.11)                          \\ \hline
Word2vec               & Naive   Bayes         & 72.37                  & 71.79               & 71.79                 & 71.31                 & (69.32,74.26)                          \\ \hline
Word2vec               & Logistic   Regression & 91.6                   & 91.9                & 91.3                  & 91.3                  & (89.75,92.84)                          \\ \hline
Word2vec               & XGBoost               & 95.54                  & \textbf{95.73}      & \textbf{95.21}        & \textbf{95.14}        & (94.04,96.39)                          \\ \hline
Word2vec               & LSTM                  & \textbf{96.08}         & 94.45               & 94.45                 & 94.45                 & (93.2,95.71)                           \\ \hline
\end{tabular}}

\end{table}

\begin{table}[htb!]
\caption[isiZulu Titles Augmented Dataset Model Performance]{isiZulu Titles Augmented Dataset Model Performance}
\label{tab:table44a}
	\centering
	\resizebox{\linewidth}{1.5cm}{
	\begin{tabular}{|c|c|c|c|c|c|c|}
\hline
\multicolumn{1}{|c|}{\textbf{Preprocessing}} & \multicolumn{1}{c|}{\textbf{Model}} & \multicolumn{1}{c|}{\textbf{Precision(\%)}} & \multicolumn{1}{c|}{\textbf{Recall(\%)}} & \multicolumn{1}{c|}{\textbf{F1-score(\%)}} & \multicolumn{1}{c|}{\textbf{Accuracy(\%)}} & \multicolumn{1}{c|}{\textbf{Confidence   Interval(f1 score)}} \\ \hline
Bag-Of-Words                              & Naive Bayes                         & 58.93                                       & 32.91                                    & 31.62                                      & 37.83                                      & (28.86,34.37)                                                 \\ \hline
Bag-Of-Words                              & Logistic Regression                 & 60.79                                       & 34.54                                    & 34.05                                      & 39.2                                       & (31.24,36.85)                                                 \\ \hline
Bag-Of-Words                              & XGBoost                             & 51.12                                       & 28.22                                    & 24.47                                      & 33.27                                      & (21.92,27.01)                                                 \\ \hline
TF-IDF                              & Naive Bayes                         & 59.45                                       & 33.25                                    & 32.3                                       & 38.1                                       & (29.54,35.07)                                                 \\ \hline
TF-IDF                              & Logistic Regression                 & 59.41                                       & 34.87                                    & 34.42                                      & 39.47                                      & (31.6,37.23)                                                  \\ \hline
TF-IDF                              & XGBoost                             & 53.33                                       & 28.85                                    & 25.41                                      & 33.82                                      & (22.83,27.98)                                                 \\ \hline
Word2vec                                     & Naive Bayes                         & 67.92                                       & 57.97                                    & 59.3                                       & 60.89                                      & (56.39,62.21)                                                 \\ \hline
Word2vec                                     & Logistic Regression                 & \textbf{86.35}                              & \textbf{87.65}                           & \textbf{86.42}                             & \textbf{85.69}                             & (84.39,88.45)                                                 \\ \hline
Word2vec                                     & XGBoost                             & 86.2                                        & 85.99                                    & 85.83                                      & 84.96                                      & (83.77,87.89)                                                 \\ \hline
Word2vec                                     & LSTM                                & 85.32                                       & 85.16                                    & 84.37                                      & 85.16                                      & (82.22,86.52)                                                 \\ \hline
\end{tabular}}

\end{table}

\begin{table}[htb!]
\caption[Siswati Titles Augmented Dataset Model Performance]{Siswati Titles Augmented Dataset Model Performance}
\label{tab:table4a}
	\centering
	\resizebox{\linewidth}{1.5cm}{
\begin{tabular}{|c|c|c|c|c|c|c|}
\hline
\multicolumn{1}{|c|}{\textbf{Preprocessing}} & \multicolumn{1}{c|}{\textbf{Model}} & \multicolumn{1}{c|}{\textbf{Precision(\%)}} & \multicolumn{1}{c|}{\textbf{Recall(\%)}} & \multicolumn{1}{c|}{\textbf{F1-score(\%)}} & \multicolumn{1}{c|}{\textbf{Accuracy(\%)}} & \multicolumn{1}{c|}{\textbf{Confidence   Interval(f1 score)}} \\ \hline
Bag-Of-Words                              & Naive Bayes                         & 71.98                                       & 69.52                                    & 69.35                                      & 68.79                                      & (61.82,76.88)                                                 \\ \hline
Bag-Of-Words                              & Logistic Regression                 & 78.78                                       & 74.8                                     & 74.74                                      & 74.31                                      & (67.65,81.84)                                                 \\ \hline
Bag-Of-Words                              & XGBoost                             & 81.99                                       & 74.7                                     & 74.47                                      & 74.33                                      & (67.35,81.59)                                                 \\ \hline
TF-IDF                              & Naive Bayes                         & 75.67                                       & 73.03                                    & 72.85                                      & 72.24                                      & (65.58,80.11)                                                 \\ \hline
TF-IDF                              & Logistic Regression                 & 78.93                                       & 75.5                                     & 75.57                                      & 75                                         & (68.55,82.59)                                                 \\ \hline
TF-IDF                              & XGBoost                             & 81.1                                        & 74.13                                    & 73.09                                      & 73.62                                      & (65.84,80.33)                                                 \\ \hline
Word2vec                                     & Naive Bayes                         & 84.26                                       & 83.41                                    & 82.52                                      & 82.66                                      & (76.32,88.73)                                                 \\ \hline
Word2vec                                     & Logistic Regression                 & 91.17                                       & 89.9                                     & 87.83                                      & 88.89                                      & (82.49,93.17)                                                 \\ \hline
Word2vec                                     & XGBoost                             & 91.57                                       & 91.33                                    & 89.8                                       & 90.22                                      & (84.86,94.74)                                                 \\ \hline
Word2vec                                     & LSTM                                & \textbf{94.88}                              & \textbf{92.41}                           & \textbf{93.15}                             & \textbf{92.41}                             & (89.02,97.27)                                                 \\ \hline
\end{tabular}}

\end{table}


\subsubsection{SMOTE}
SMOTE is an oversampling technique used to rebalance the original training set through the creation of synthetic samples of the minority class \cite{fernandez2018smote}. This technique works by selecting the minority class and the total amount of oversampling to balance the classes, then the k-nearest neighbours for that particular class are obtained , therefore, iteratively the k nearest neighbours are randomly chosen to create new instances \cite{fernandez2018smote}. This oversampling technique was used to balance the classes and increase the dataset. Note that SMOTE uses a different approach from the Data Augmentation approach presented earlier.

We applied SMOTE on our three datasets and run the classification model using 5-fold cross validation, the results from each dataset are presented below. From the below tables, it was observed that Word2vec produced the best classification models from all the three datasets. XGBoost performed well in all instances scoring f1-score of 93.35\%, 91.26\%, 87.46\% for isiZulu articles, isiZulu titles and Siswati titles datasets respectively. We observed the XGBoost model on isiZulu articles struggled to separate \textit{society} and \textit{politics} from \textit{crime,law and justice} since most of the incorrect classification happened in the instance where \textit{society} and \textit{politics} were classified as \textit{crime,law and justice}.

\begin{table}[htb!]
\caption[isiZulu Articles SMOTE Dataset Model Performance]{isiZulu Articles SMOTE Dataset Model Performance}
\label{tab:table4b}
	\centering
	\resizebox{\linewidth}{1.5cm}{
	\begin{tabular}{|c|c|c|c|c|c|c|}
\hline
\textbf{Preprocessing} & \textbf{Model}        & \textbf{Precision(\%)} & \textbf{Recall(\%)} & \textbf{F1-score(\%)} & \textbf{Accuracy(\%)} & \textbf{Confidence Interval(f1 score)} \\ \hline
Bag-Of-Words        & Naive   Bayes         & 56.37                  & 39.06               & 36.63                 & 39.04                 & (34.16,39.11)                          \\ \hline
Bag-Of-Words        & Logistic   Regression & 55.67                  & 51.19               & 50.08                 & 51.16                 & (47.52,52.65)                          \\ \hline
Bag-Of-Words        & XGBoost               & 82.31                  & 76.34               & 75.99                 & 76.37                 & (73.8,78.18)                           \\ \hline
TF-IDF        & Naive   Bayes         & 78.93                  & 77.81               & 76.83                 & 77.81                 & (74.67,79.0)                           \\ \hline
TF-IDF        & Logistic   Regression & 82.2                   & 82.38               & 81.68                 & 82.4                  & (79.7,83.66)                           \\ \hline
TF-IDF        & XGBoost               & 81.7                   & 79.17               & 79.51                 & 79.18                 & (77.44,81.58)                          \\ \hline
Word2vec               & Naive   Bayes         & 74.44                  & 74.25               & 74.12                 & 74.25                 & (71.87,76.37)                          \\ \hline
Word2vec               & Logistic   Regression & 92.43                  & 92.11               & 91.88                 & 92.12                 & (90.48,93.28)                          \\ \hline
Word2vec               & XGBoost               & \textbf{93.75}         & \textbf{93.55}      & \textbf{93.35}        & \textbf{93.56}        & (92.08,94.63)                          \\ \hline

\end{tabular}}

\end{table}
\begin{table}[htb!]
\caption[isiZulu Titles SMOTE Dataset Model Performance]{isiZulu Titles SMOTE Dataset Model Performance}
\label{tab:table4b1}
	\centering
	\resizebox{\linewidth}{1.5cm}{
	    \begin{tabular}{|c|c|c|c|c|c|c|}
\hline
\textbf{Preprocessing} & \textbf{Model}        & \textbf{Precision(\%)} & \textbf{Recall(\%)} & \textbf{F1-score(\%)} & \textbf{Accuracy(\%)} & \textbf{Confidence Interval(f1 score)} \\ \hline
Bag-Of-Words        & Naive   Bayes         & 36.92                  & 23.33               & 15.91                 & 23.22                 & (14.03,17.78)                          \\ \hline
Bag-Of-Words        & Logistic   Regression & 46.08                  & 25.9                & 18.23                 & 25.89                 & (16.25,20.21)                          \\ \hline
Bag-Of-Words        & XGBoost               & 65.14                  & 38.34               & 37.52                 & 38.36                 & (35.03,40.0)                           \\ \hline
TF-IDF        & Naive   Bayes         & 64.37                  & 37.69               & 37.38                 & 37.6                  & (34.9,39.86)                           \\ \hline
TF-IDF        & Logistic   Regression & 65.23                  & 39.72               & 39.71                 & 39.73                 & (37.2,42.22)                           \\ \hline
TF-IDF        & XGBoost               & 65.6                   & 38.2                & 37.56                 & 38.22                 & (35.08,40.05)                          \\ \hline
Word2vec               & Naive   Bayes         & 74.49                  & 74.02               & 73.85                 & 74.04                 & (71.6,76.11)                           \\ \hline
Word2vec               & Logistic   Regression & 91.56                  & 91.08               & 90.63                 & 91.1                  & (89.13,92.12)                          \\ \hline
Word2vec               & XGBoost               & 91.96                  & 91.56               & 91.26                 & 91.58                 & (89.81,92.71)                          \\ \hline
Word2vec               & LSTM                  & 73.53                  & 72.82               & 72.75                 & 72.82                 & (69.08,76.43)                          \\ \hline
\end{tabular}}

\end{table}

\begin{table}[htb!]
\caption[Siswati Titles SMOTE Dataset Model Performance]{Siswati Titles SMOTE Dataset Model Performance}
\label{tab:table41b}
	\centering
	\resizebox{\linewidth}{1.5cm}{

	\begin{tabular}{|c|c|c|c|c|c|c|}
\hline
\multicolumn{1}{|c|}{\textbf{Preprocessing}} & \multicolumn{1}{c|}{\textbf{Model}} & \multicolumn{1}{c|}{\textbf{Precision(\%)}} & \multicolumn{1}{c|}{\textbf{Recall(\%)}} & \multicolumn{1}{c|}{\textbf{F1-score(\%)}} & \multicolumn{1}{c|}{\textbf{Accuracy(\%)}} & \multicolumn{1}{c|}{\textbf{Confidence   Interval(f1 score)}} \\ \hline
Bag-Of-Words                              & Naive Bayes                         & 60.63                                       & 40.67                                    & 37.91                                      & 40                                         & (30.4,45.43)                                                  \\ \hline
Bag-Of-Words                              & Logistic Regression                 & 65.03                                       & 44.19                                    & 42.91                                      & 44.38                                      & (35.24,50.58)                                                 \\ \hline
Bag-Of-Words                              & XGBoost                             & 81.3                                        & 74.38                                    & 73.65                                      & 74.38                                      & (66.83,80.48)                                                 \\ \hline
TF-IDF                              & Naive Bayes                         & 80.71                                       & 79.14                                    & 74.32                                      & 78.75                                      & (67.55,81.09)                                                 \\ \hline
TF-IDF                              & Logistic Regression                 & 82.25                                       & 82.95                                    & 80.42                                      & 83.12                                      & (74.27,86.57)                                                 \\ \hline
TF-IDF                              & XGBoost                             & 85.47                                       & 77.05                                    & 76.6                                       & 76.88                                      & (70.04,83.16)                                                 \\ \hline
Word2vec                                     & Naive Bayes                         & 85.86                                       & 83.71                                    & 82.5                                       & 83.75                                      & (76.62,88.39)                                                 \\ \hline
Word2vec                                     & Logistic Regression                 & \textbf{90.35}                              & 88.1                                     & 86.2                                       & 88.12                                      & (80.86,91.55)                                                 \\ \hline
Word2vec                                     & XGBoost                             & 89.88                                       & \textbf{88.76}                           & \textbf{87.46}                             & \textbf{88.75}                             & (82.33,92.59)                                                 \\ \hline
\end{tabular}}

\end{table}


\section{Summary}

We observed that Data Augmentation outperformed SMOTE in two instances, that is, isiZulu articles and Siswati titles datasets, whereas SMOTE outperformed Data augmentation only in case of isiZulu titles dataset, however, we hope to look into the difference performance from these re-sampling techniques and have a confirmatory pipeline to provide guidance on what approach to take under what circumstance. However, we present the generalised pipeline obtained from this work as a baseline.

\begin{figure}[h]
	\centering {\includegraphics[width=0.425\textwidth]{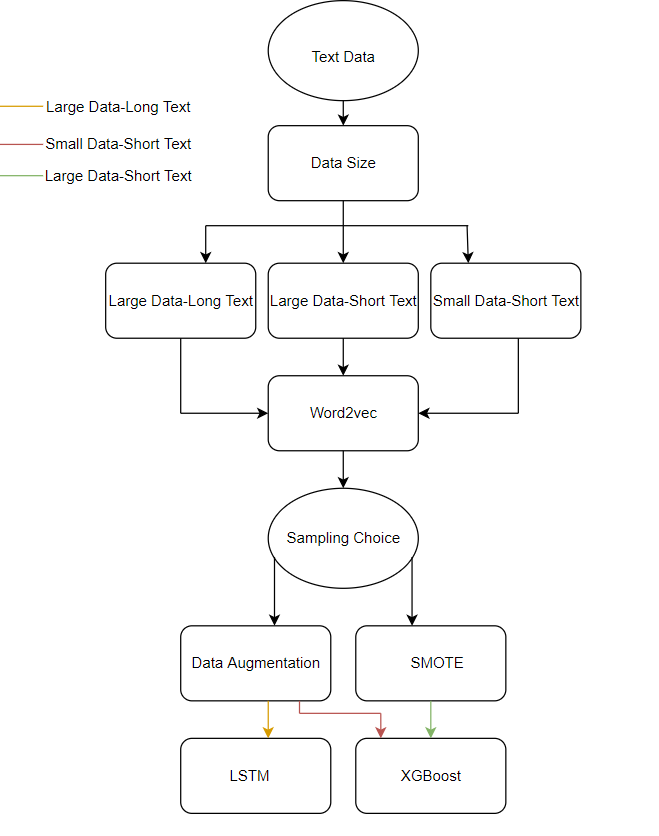}}
	\caption[Recommended Pipeline]{Recommended Pipeline}
	\label{fig:figta31a}
\end{figure}

The Pipeline obtained from this work was summarised and presented in  figure ~\ref{fig:figta31a} below together with the corresponding top performing classification models presented in table ~\ref{fig:figta31a}, the figure ~\ref{fig:figta31a} shows the choices that produced the best results under different circumstances for three different datasets. It was observed that the datasets used resembled three different qualities, that is, large size and long-text (isiZulu Articles), large size and short text(isiZulu Titles), and small size  and short text(Siswati), these varieties produced different outcomes from the models under the same circumstance and can be generalised as follows:
\begin{itemize}
    \item If the data size is large and contains long-text then Contextual Data Augmentation is recommended over SMOTE, and LSTM is likely to perform better.
    \item If the data size is large and contains short-text then SMOTE is recommended over Contextual Data Augmentation, and XGBoost is likely to perform better.
    
    \item If the data size is small and contains short-text then Contextual Data Augmentation is recommended over SMOTE, and XGBoost is likely to perform better
\end{itemize}

The Above generalisation is limited to Word2vec word embedding since it is the one that produced outstanding results from all the datasets as compared to TFIDF and Bag-Of-Words. It remains a task to further investigate the poor performance from TFIDF and Bag-Of-Words, possibly the parameter change in classification could lead to good results.

\begin{table}[htb!]
\caption[Top Performing Classification Models]{Top Performing Classification Models}
\label{tab:fig31a0}
	\centering 
	\resizebox{\linewidth}{1.5cm}{
	\begin{tabular}{|c|c|c|c|c|c|c|c|c|c|}
\hline
\multicolumn{9}{|c|}{\textbf{Best Model based on Sampling technique}}                                                                                                                                                                                                                                                                                                                                        \\ \hline
\multicolumn{1}{|c|}{\textbf{Dataset}} & \multicolumn{1}{c|}{\textbf{Sampling}} & \multicolumn{1}{c|}{\textbf{Word embbeding}} & \multicolumn{1}{c|}{\textbf{Model}}      & \multicolumn{1}{c|}{\textbf{Precision(\%)}} & \multicolumn{1}{c|}{\textbf{Recall(\%)}} & \multicolumn{1}{c|}{\textbf{F1-score(\%)}} & \multicolumn{1}{c|}{\textbf{Accuracy(\%)}} & \textbf{Confidence   Interval(f1 score)} \\ \hline
\multicolumn{1}{|c|}{isiZulu Articles} & \multicolumn{1}{c|}{Augmented}         & \multicolumn{1}{c|}{Word2vec}                & \multicolumn{1}{c|}{XGBoost}             & \multicolumn{1}{c|}{95.54}                  & \multicolumn{1}{c|}{95.73}               & \multicolumn{1}{c|}{95.21}                 & \multicolumn{1}{c|}{95.14}                 & (94.04,96.39)                            \\ \hline
\multicolumn{1}{|c|}{isiZulu Titles}   & \multicolumn{1}{c|}{Augmented}         & \multicolumn{1}{c|}{Word2vec}                & \multicolumn{1}{c|}{Logistic Regression} & \multicolumn{1}{c|}{86.35}                  & \multicolumn{1}{c|}{87.65}               & \multicolumn{1}{c|}{86.42}                 & \multicolumn{1}{c|}{85.69}                 & (84.39,88.45)                            \\ \hline
\multicolumn{1}{|c|}{Siswati Titles}   & \multicolumn{1}{c|}{Augmented}         & \multicolumn{1}{c|}{Word2vec}                & \multicolumn{1}{c|}{LSTM}                & \multicolumn{1}{c|}{94.88}                  & \multicolumn{1}{c|}{92.41}               & \multicolumn{1}{c|}{93.15}                 & \multicolumn{1}{c|}{92.41}                 & (89.02,97.27)                            \\ \hline
\multicolumn{1}{|c|}{}                 & \multicolumn{1}{c|}{}                  & \multicolumn{1}{c|}{}                        & \multicolumn{1}{c|}{}                    & \multicolumn{1}{c|}{}                       & \multicolumn{1}{c|}{}                    & \multicolumn{1}{c|}{}                      & \multicolumn{1}{c|}{}                      &                                          \\ \hline
\multicolumn{1}{|c|}{isiZulu Articles} & \multicolumn{1}{c|}{SMOTE}             & \multicolumn{1}{c|}{Word2vec}                & \multicolumn{1}{c|}{XGBoost}             & \multicolumn{1}{c|}{93.75}                  & \multicolumn{1}{c|}{93.55}               & \multicolumn{1}{c|}{93.35}                 & \multicolumn{1}{c|}{93.56}                 & (92.08,94.63)                            \\ \hline
\multicolumn{1}{|c|}{isiZulu Titles}   & \multicolumn{1}{c|}{SMOTE}             & \multicolumn{1}{c|}{Word2vec}                & \multicolumn{1}{c|}{XGBoost}             & \multicolumn{1}{c|}{91.96}                  & \multicolumn{1}{c|}{91.56}               & \multicolumn{1}{c|}{91.26}                 & \multicolumn{1}{c|}{91.58}                 & (89.81,92.71)                            \\ \hline
\multicolumn{1}{|c|}{Siswati Titles}   & \multicolumn{1}{c|}{SMOTE}             & \multicolumn{1}{c|}{Word2vec}                & \multicolumn{1}{c|}{XGBoost}             & \multicolumn{1}{c|}{89.88}                  & \multicolumn{1}{c|}{88.76}               & \multicolumn{1}{c|}{87.46}                 & \multicolumn{1}{c|}{88.75}                 & (82.33,92.59)                            \\ \hline
\end{tabular}}

\end{table}

\section{Conclusion and Future Work}
This work introduced the collection and annotation of isiZulu and Siswati news datasets. There is still a data shortage (more especially annotated data) of these two native languages, especially Siswati. However, this work paved a way for the other researchers who would want to use annotated data for isiZulu and/or Siswati in downstream NLP tasks.

The experimental findings from the classification models and different combinations of word embeddings with model baselines were presented. Though we were limited by the data availability, however, this provides an overview of what could be achieved with minimal datasets. The isiZulu and Siswati annotated datasets will be made available for other researchers, the pre-trained vectorizers will be open-sourced to other researchers and the classification results that maybe be used as benchmarks.

The collection and annotation of native language datasets remain a task for the future. For this to be successful, there needs to be an identification of other language sources where the dataset can be extracted for more models to be trained. Furthermore, NLP researchers need to focus more on effective ways to augment the datasets. They should be compared with SMOTE sampling, because of the imbalance in the dataset. It is beneficial to have effective ways to augment native language datasets. 

In addition, it is also worth investigating the poor performance of TFIDF and Bag-Of-Words compared to Word2vec, possible investigation areas could be the word embedding nature and the classification models hyperparameters optimisation that could improve classification performance. Another extension of this work is transfer learning from isiZulu to Siswati. The isiZulu dataset is large compared to the Siswati dataset making it a viable avenue of research to investigate if transfer learning improves the classification performance for Siswati in this context.

\printendnotes[custom]


\bibliographystyle{agsm}

\bibliography{bibliography.bib}

@misc{artificialintelligencebigdataanalyticsandinsight_2020, title={What is NLP and Why is it Important?},
author={Priya Dialani},url={https://www.analyticsinsight.net/what-is-nlp-and-why-is-it-important/\#:$\sim$:text=Natural language\%2}, journal={Artificial Intelligence, Big Data Analytics and Insight}, year={2020}, month={May}}

@inproceedings{duvenhage2017improved,
  title={Improved text language identification for the South African languages},
  author={Duvenhage, Bernardt and Ntini, Mfundo and Ramonyai, Phala},
  booktitle={2017 Pattern Recognition Association of South Africa and Robotics and Mechatronics (PRASA-RobMech)},
  pages={214--218},
  year={2017},
  organization={IEEE}
}

@misc{gateway_2021, 
author={Mary Alexander},
title={The 11 languages of South Africa}, url={https://southafrica-info.com/arts-culture/11-languages-south-africa/}, journal={South Africa Gateway},  year={2021}, month={May}}

@article{kobayashi2018contextual,
  title={Contextual augmentation: Data augmentation by words with paradigmatic relations},
  author={Kobayashi, Sosuke},
  journal={arXiv preprint arXiv:1805.06201},
  year={2018}
}

@article{zhang2015character,
  title={Character-level convolutional networks for text classification},
  author={Zhang, Xiang and Zhao, Junbo and LeCun, Yann},
  journal={Advances in neural information processing systems},
  volume={28},
  pages={649--657},
  year={2015}
}

@article{oh2020time,
  title={Time-series Data Augmentation based on Interpolation},
  author={Oh, Cheolhwan and Han, Seungmin and Jeong, Jongpil},
  journal={Procedia Computer Science},
  volume={175},
  pages={64--71},
  year={2020},
  publisher={Elsevier}
}

@inproceedings{abonizio2020pre,
  title={Pre-trained Data Augmentation for Text Classification},
  author={Abonizio, Hugo Queiroz and Junior, Sylvio Barbon},
  booktitle={Brazilian Conference on Intelligent Systems},
  pages={551--565},
  year={2020},
  organization={Springer}
}

@inproceedings{baumann2014using,
  title={Using resource-rich languages to improve morphological analysis of under-resourced languages},
  author={Baumann, Peter and Pierrehumbert, Janet},
  booktitle={Proceedings of the Ninth International Conference on Language Resources and Evaluation (LREC'14)},
  pages={3355--3359},
  year={2014}
}

@inproceedings{bonab2019simulating,
  title={Simulating CLIR translation resource scarcity using high-resource languages},
  author={Bonab, Hamed and Allan, James and Sitaraman, Ramesh},
  booktitle={Proceedings of the 2019 ACM SIGIR International Conference on Theory of Information Retrieval},
  pages={129--136},
  year={2019}
}

@article{bosch2008experimental,
  title={Experimental bootstrapping of morphological analysers for Nguni languages},
  author={Bosch, Sonja and Pretorius, Laurette and Fleisch, Axel},
  journal={Nordic Journal of African Studies},
  volume={17},
  number={2},
  pages={23--23},
  year={2008}
}

@article{duong2021review,
  title={A review: preprocessing techniques and data augmentation for sentiment analysis},
  author={Duong, Huu-Thanh and Nguyen-Thi, Tram-Anh},
  journal={Computational Social Networks},
  volume={8},
  number={1},
  pages={1--16},
  year={2021},
  publisher={SpringerOpen}
}

@inproceedings{eiselen2014developing,
  title={Developing Text Resources for Ten South African Languages.},
  author={Eiselen, Roald and Puttkammer, Martin J},
  booktitle={LREC},
  pages={3698--3703},
  year={2014}
}

@article{fang2017model,
  title={Model transfer for tagging low-resource languages using a bilingual dictionary},
  author={Fang, Meng and Cohn, Trevor},
  journal={arXiv preprint arXiv:1705.00424},
  year={2017}
}

@inproceedings{hsueh2009data,
  title={Data quality from crowdsourcing: a study of annotation selection criteria},
  author={Hsueh, Pei-Yun and Melville, Prem and Sindhwani, Vikas},
  booktitle={Proceedings of the NAACL HLT 2009 workshop on active learning for natural language processing},
  pages={27--35},
  year={2009}
}

@article{marivate2020investigating,
  title={Investigating an approach for low resource language dataset creation, curation and classification: Setswana and Sepedi},
  author={Marivate, Vukosi and Sefara, Tshephisho and Chabalala, Vongani and Makhaya, Keamogetswe and Mokgonyane, Tumisho and Mokoena, Rethabile and Modupe, Abiodun},
  journal={arXiv preprint arXiv:2003.04986},
  year={2020}
}

@article{nguyen2017transfer,
  title={Transfer learning across low-resource, related languages for neural machine translation},
  author={Nguyen, Toan Q and Chiang, David},
  journal={arXiv preprint arXiv:1708.09803},
  year={2017}
}

@article{niyongabo2020kinnews,
  title={KINNEWS and KIRNEWS: Benchmarking cross-lingual text classification for Kinyarwanda and Kirundi},
  author={Niyongabo, Rubungo Andre and Qu, Hong and Kreutzer, Julia and Huang, Li},
  journal={arXiv preprint arXiv:2010.12174},
  year={2020}
}

@inproceedings{rakholia2016lexical,
  title={Lexical classes based stop words categorization for Gujarati language},
  author={Rakholia, Rajnish M and Saini, Jatinderkumar R},
  booktitle={2016 2nd international conference on advances in computing, communication, \& automation (ICACCA)(Fall)},
  pages={1--5},
  year={2016},
  organization={IEEE}
}

@article{shamsfardchallenges,
  title={Challenges and Opportunities in Processing Low Resource Languages: A study on Persian},
  author={Shamsfard, Mehrnoush}
}

@inproceedings{stenetorp2012brat,
  title={BRAT: a web-based tool for NLP-assisted text annotation},
  author={Stenetorp, Pontus and Pyysalo, Sampo and Topi{\'c}, Goran and Ohta, Tomoko and Ananiadou, Sophia and Tsujii, Jun’ichi},
  booktitle={Proceedings of the Demonstrations at the 13th Conference of the European Chapter of the Association for Computational Linguistics},
  pages={102--107},
  year={2012}
}

@article{tang2018improving,
  title={Improving multilingual semantic textual similarity with shared sentence encoder for low-resource languages},
  author={Tang, Xin and Cheng, Shanbo and Do, Loc and Min, Zhiyu and Ji, Feng and Yu, Heng and Zhang, Ji and Chen, Haiqin},
  journal={arXiv preprint arXiv:1810.08740},
  year={2018}
}

@misc{whyatt2019languages,
  title={Languages of low diffusion and low resources: translation research and training challenges: Special Issue Proposal ITT--15 (1), March 2021},
  author={Whyatt, Bogus{\l}awa and Pavlovi{\'c}, Nata{\v{s}}a},
  year={2019},
  publisher={Taylor \& Francis}
}

@article{xu2013cross,
  title={Cross-lingual language modeling for low-resource speech recognition},
  author={Xu, Ping and Fung, Pascale},
  journal={IEEE transactions on audio, speech, and language processing},
  volume={21},
  number={6},
  pages={1134--1144},
  year={2013},
  publisher={IEEE}
}

@article{zoph2016transfer,
  title={Transfer learning for low-resource neural machine translation},
  author={Zoph, Barret and Yuret, Deniz and May, Jonathan and Knight, Kevin},
  journal={arXiv preprint arXiv:1604.02201},
  year={2016}
}

@article{mesham2021low,
  title={Low-Resource Language Modelling of South African Languages},
  author={Mesham, Stuart and Hayward, Luc and Shapiro, Jared and Buys, Jan},
  journal={arXiv preprint arXiv:2104.00772},
  year={2021}
}

@article{nyoni2021low,
  title={Low-Resource Neural Machine Translation for Southern African Languages},
  author={Nyoni, Evander and Bassett, Bruce A},
  journal={arXiv preprint arXiv:2104.00366},
  year={2021}
}

@article{fernandez2018smote,
  title={SMOTE for learning from imbalanced data: progress and challenges, marking the 15-year anniversary},
  author={Fern{\'a}ndez, Alberto and Garcia, Salvador and Herrera, Francisco and Chawla, Nitesh V},
  journal={Journal of artificial intelligence research},
  volume={61},
  pages={863--905},
  year={2018}
}

@inproceedings{NIPS2013_9aa42b31,
 author = {Mikolov, Tomas and Sutskever, Ilya and Chen, Kai and Corrado, Greg S and Dean, Jeff},
 booktitle = {Advances in Neural Information Processing Systems},
 editor = {C.J. Burges and L. Bottou and M. Welling and Z. Ghahramani and K.Q. Weinberger},
 pages = {},
 publisher = {Curran Associates, Inc.},
 title = {Distributed Representations of Words and Phrases and their Compositionality},
 url = {https://proceedings.neurips.cc/paper/2013/file/
 9aa42b31882ec039965f3c4923ce901b-Paper.pdf},
 volume = {26},
 year = {2013}
}

\end{document}